# Discriminative Autoencoder for Feature Extraction: Application to Character Recognition

Anupriya Gogna and Angshul Majumdar

Indraprastha Institute of Information Technology

anupriyag@iiitd.ac.in and angshul@iiitd.ac.in

Abstract –

Conventionally, autoencoders are unsupervised representation learning tools. In this work, we propose a novel discriminative autoencoder. Use of supervised discriminative learning ensures that the learned representation is robust to variations commonly encountered in image datasets. Using the basic discriminating autoencoder as a unit, we build a stacked architecture aimed at extracting relevant representation from the training data. The efficiency of our feature extraction algorithm ensures a high classification accuracy with even simple classification schemes like KNN (K-nearest neighbor). We demonstrate the superiority of our model for representation learning by conducting experiments on standard datasets for character/image recognition and subsequent comparison with existing supervised deep architectures like class sparse stacked autoencoder and discriminative deep belief network.

Keywords - Autoencoder, feature extraction, classification, supervised learning

## 1. Introduction

Automatic classification of images assumes great relevance owing to the large size of the database required to be processed in various application areas, ranging from hyperspectral imaging [1] to biomedical analysis [2].

One of the prime requirements for effective classification is the selection of relevant features adept at capturing internal structure of the data [3]. In addition, an effective feature set should display high inter-class diversity and adequate robustness to variations such as illumination effects, rotation and translation. Also, to reduce the computational burden on the classification algorithm, a reduced feature length compared to the raw data, is desired.

Traditionally, there are two classes of feature extraction techniques. 1) Handcrafted features like Histogram of Oriented Gradients (HoG) [4], Scale Invariant Feature Transform (SIFT) [5], Local Binary Pattern (LBP) [6] have enjoyed significant popularity for various image-based classification tasks. These features require deep insight into imaging. 2) On the other hand, statistical feature extraction techniques like Eigenface [7], Fisherface [8], local Principal Component Analysis (2D-PCA) [9] etc. are more abstract and are based on optimizing some mathematical costs.

In recent years, representation learning, which includes both deep learning (stacked autoencoder and deep belief network) and dictionary learning, has gained immense popularity owing to their astounding success in various fields including image classification [10] and speech processing [11] amongst others. These are automated feature extraction techniques that require little insight into the problem.

In recent times, several works [12], [13] have motivated the use of deep networks to effectively learn abstract features in a hierarchical fashion. These deep architectures have been shown to yield more robust and comprehensive representation [14] of input data for classification [15].

The basic building blocks of these deep architectures are either the stochastic RBM's (Restricted Boltzmann Machines) [16] or the deterministic Autoencoders [17]. Given a training dataset, RBM tries to learn the network weights such that the similarity between the projection (of the training data) and the learned representation is maximized. Autoencoders (AE) on the other hand consists of two networks. The first one maps the input (training data) to the representation / feature space; the second network maps the representation space to the output (training data). Thus, an AE approximates an Identity operator; which may sound trivial, but by constraining the nodes or connections of the networks one can learn interesting representations of the data.

RBM and AE are shallow architectures. Proponents of deep learning believe that better (compact / abstract) representation can be learnt by going deeper. However, learning the network weights for several layers is a difficult task. Usually, there is not enough data, the network overfits and loses its generalization ability thereby yielding subpar results at operational stage. In [17], authors presented a greedy mechanism to train the multilayer (stacked) architectures wherein each of the layer is individually trained to yield best possible representation which in turn acts as input to subsequent layer. Greedy approach learns only one network at a time, it has fewer parameters to learn, so even with limited training data, it yields better results during operation.

In this work, we modify the basic (unsupervised) autoencoder by introducing a discriminative penalty in the cost function itself. In general, it has been found that better results are attained with supervised (discriminative) learning tools compared to unsupervised ones. Following prior studies on deep learning, especially on stacked autoencoders, we use our discriminative autoencoder as a basic unit to form deeper architectures.

An autoencoder [17] is a self-supervised neural network, i.e. input and output are the same. It is unsupervised in the sense that the training does not require any class information. It consists of two parts – the encoder maps the input to a latent space, and the decoder maps the latent representation to the data. For a given input vector (including the bias term) $\mathbf{x}$, the latent space is expressed as#:

$$\mathbf{h} = \phi(\mathbf{W_{i\text{-}h}x}) \quad (1)$$

Here the rows of $\mathbf{W_{i\text{-}h}}$ are the link weights from all the input nodes to the corresponding latent node. The activation function $\phi(\bullet)$ can be linear [18, 19], but in most cases it is non-linear (sigmoid, tanh etc.)

The decoder portion reverse maps the latent features to the data space as in (2). In (1) and (2), subscripts i, h, and o stand for input, hidden and output layers.

$$\mathbf{x} = \mathbf{W_{h\text{-}o}}\phi(\mathbf{W_{i\text{-}h}x}) \quad (2)$$

Since the data space is assumed to be the space of real numbers, there is no sigmoidal function here.

During training, the problem is to learn the encoding and decoding weights – $\mathbf{W_{i\text{-}h}}$ and $\mathbf{W_{h\text{-}o}}$. This is achieved by minimizing the Euclidean cost:

$$\underset{\mathbf{W_{h\text{-}o}},\mathbf{W_{i\text{-}h}}}{\arg\min} \left\| \mathbf{X} - \mathbf{W_{h\text{-}o}}\phi(\mathbf{W_{i\text{-}h}X}) \right\|_F^2 \quad (3)$$

Here $\mathbf{X} = [\mathbf{x_1} \mid ... \mid \mathbf{x_N}]$ consists of all the training samples stacked as columns of the matrix. The problem in (3) is clearly non-convex. However, it is solved easily by gradient descent techniques since the activation function is smooth and continuously differentiable.

In this work, we present a novel discriminative autoencoder (DiAE). We augment the standard $l_2$-norm loss function (3) with additional regularization constraints derived from available class information. Our design enforces the derived feature vectors to be consistent with the corresponding class labels via a linear mapping. Supervised learning of hidden layer variables ascertains that features corresponding to images of a class are mapped to the (same) class label via a fixed (but derived) linear mapping. Thus, it mitigates the impact of any variations in images, related to illumination, rotation, background information and others, owing to a more robust design.

Stacking of AE [20] to construct a multilayer architecture has been shown to yield better performance than standalone shallow units for classification. Thus, we use our DiAE as the building block of a deep (multilayer) feature extractor. We adopt the greedy scheme;

training each layer independently and using the representation / features as input to next stage. Each layer is expected to give a more abstract representation of the input data compared to the previous layer; representation from the final layer is used as input to a standard classifier.

We conduct experiments on the various OCR (optical character recognition) datasets to showcase our model's capability to generate more robust and discriminatory features compared to a standard AE.

## 1.1. Related Work

Motivated by studies on neural activity [21], several works promoted recovery of an over complete but sparse hidden layer [22], [23]. For this, they added a regularization term to (3) which penalizes any deviation of the feature vector from the desired sparse behavior. Such a constraint, though gives superior performance than standard AE, does not give explicit dimensionality reduction – an important component of an effective feature extraction scheme.

Stacked denoising autoencoders (SDAE) [20] are a variant of the basic autoencoder where the input consists of noisy samples and the output consists of clean samples; this is a stochastic regularization technique. Here the encoder and decoder are learnt to denoise noisy input samples. The learned features appear to be more robust when learnt by SDAE compared to standard stacked AE.

In a recent work a Marginalized Denoising autoencoder was proposed [24]; it does not learn a representation but learns the mapping from the input to the output. Such an autoencoder cannot be used for representation learning and associated problems but can be used for domain adaptation.

There are few prior works which propose modifications to standard AE design to improve inter-class discrimination amongst features. Authors in [25] introduced a discriminative AE by combining the HOG (Histogram of Gradients) based feature selection with manifold learning. Their discriminative AE structure works on top of linear SVM classifier built using HOG features; thus requiring as many discriminative AE as the number of SVM classifiers in stage 1. In addition to being a complex design, use of HOG in [25] requires heuristic parameter selection (such as window size) and deep insight into the image. Our model, being fully automated and based on representation learning, requires no knowledge of image structure. In [26] a discriminative non-negative matrix factorization framework is proposed. It essentially learns a basis (dictionary) matrix and its associated coefficient matrix for classification task by introducing within-class and between-class variation penalty terms in the base formulation. Their procedure requires solving an optimization problem for both the training as well as test phase which, as reported in their work itself, is more time consuming than similar existing works. We develop a discriminative AE module, unlike the dictionary learning employed in the former, wherein the training procedure can be carried out offline and during test phase a simple distance measure needs to be computed for classification. Also, our design ensures much shorter training times as compared to existing AE modules.

# 2. Proposed Approach

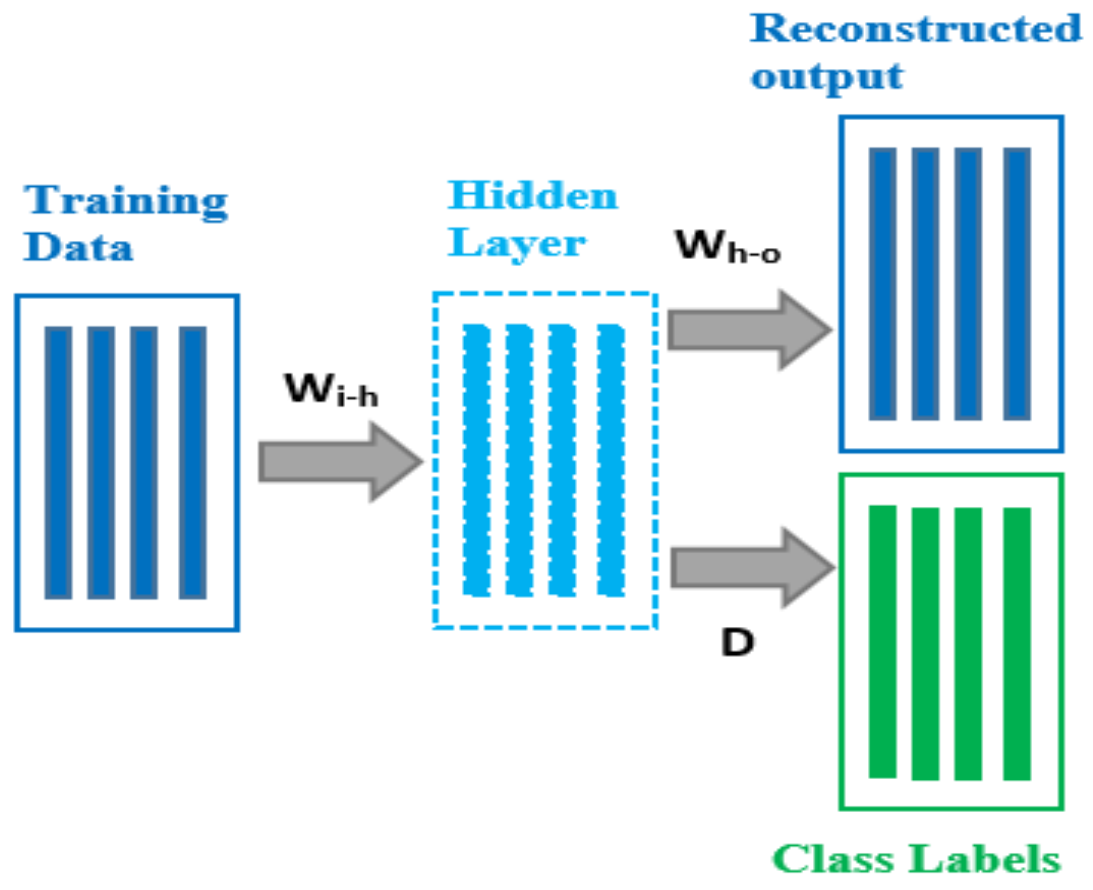

Fig. 1. Proposed Discriminative Autoencoder (DiAE)

Our proposed design for discriminative AE is shown in fig 1. Our formulation is motivated by prior studies in discriminative [27] or label-consistent [28] dictionary learning and discriminative restricted Boltzmann machine [29]. In all these studies, a linear map was learnt from the representation layer to the class labels. Such a discriminative penalty has not been imposed on autoencoder learning before. This is the first work to propose such a discriminative penalty; our formulation is as follows.

$$\underset{\mathbf{W_{h\text{-}o}},\mathbf{W_{i\text{-}h}},\mathbf{D}}{\arg\min} \left\|\mathbf{X}-\mathbf{W_{h\text{-}o}}\phi(\mathbf{W_{i\text{-}h}X})\right\|_F^2 + \lambda\left\|\mathbf{L}-\mathbf{D}\phi(\mathbf{W_{i\text{-}h}X})\right\|_F^2 \quad (4)$$

Here L is the class labels (having ones in the position of the correct class and zeroes elsewhere, D is the linear map, we will learn).

In (4), the first term preserves information from the input/output to the representation - effectively ensuring that the feature vectors display a large correlation with the training data (standard autoencoder cost function). The second, discriminative term, promotes a solution of encoder weights such that the features for same class inputs map to same class labels under a fixed (not predetermined) linear map.

## 2.1. Derivation

Our objective is to solve (4). We substitute, the latent representation as $\mathbf{Z}=\phi(\mathbf{W_{i\text{-}h}X})$ ; this recasts (4) as follows,

$$\underset{\mathbf{W_{h\text{-}o}},\mathbf{W_{i\text{-}h}},\mathbf{Z},\mathbf{D}}{\arg\min} \left\|\mathbf{X}-\mathbf{W_{h\text{-}o}Z}\right\|_F^2 + \lambda\left\|\mathbf{L}-\mathbf{DZ}\right\|_F^2 \text{ such that } \mathbf{Z}=\phi(\mathbf{W_{i\text{-}h}X}) \quad (5)$$

We can formulate the Lagrangian from (5); however the Lagrangian will impose equality between the variable (W) and the corresponding proxy (Z) in every iteration. This is not required in practice. We only want the two to be equal at convergence. Therefore instead of the Lagrangian we form the augmented Lagrangian.

$$\underset{\mathbf{W_{h\text{-}o}},\mathbf{W_{i\text{-}h}},\mathbf{Z},\mathbf{D}}{\arg\min} \left\|\mathbf{X}-\mathbf{W_{h\text{-}o}Z}\right\|_F^2 + \lambda\left\|\mathbf{L}-\mathbf{DZ}\right\|_F^2 + \mu\left\|\mathbf{Z}-\phi(\mathbf{W_{i\text{-}h}X})\right\|_F^2 \quad (6)$$

For small values of μ the equality constraint is relaxed and for large values it is enforced. One heuristic way to solve the problem would be to start with a small value of μ, solve (6); increase the value of μ, solve (6) again and keep repeating. However, this is not an elegant solution. Also, one needs to rely on intuition for increasing the value of μ. A better approach is to introduce a Bregman variable between the proxy and the original variable [30]. The Bregman variable can be automatically updated, keeping the value of μ fixed. The update of the Bregman variable would ensure that the proxy and the variable are equal during convergence.

This leads to our final formulation (7), where B is the Bregman variable.

$$\underset{\mathbf{W_{h\text{-}o}},\mathbf{W_{i\text{-}h}},\mathbf{Z},\mathbf{D}}{\arg\min} \left\|\mathbf{X}-\mathbf{W_{h\text{-}o}Z}\right\|_F^2 + \lambda\left\|\mathbf{L}-\mathbf{DZ}\right\|_F^2 + \mu\left\|\mathbf{Z}-\phi(\mathbf{W_{i\text{-}h}X})-\mathbf{B}\right\|_F^2 \quad (7)$$

We have introduced an auxiliary variable. But since it is auxiliary, it is not independent hence estimation of Z does not add to the woes of over-fitting.

Using alternating direction method of multipliers, we can segregate (7) into the following sub-problems. The idea here in is to update each of the variables separately assuming the others to be constant.

$$\text{P1}: \underset{\mathbf{W_{h\text{-}o}}}{\arg\min} \left\|\mathbf{X}-\mathbf{W_{h\text{-}o}Z}\right\|_F^2$$

$$\text{P2}: \underset{D}{\arg\min} \left\|\mathbf{L}-\mathbf{DZ}\right\|_F^2$$

P3: $\arg\min_{\mathbf{W_{i\text{-}h}}} \|\mathbf{Z}-\phi(\mathbf{W_{i\text{-}h}X})-\mathbf{B}\|_F^2 \equiv \|\phi^{-1}(\mathbf{Z}-\mathbf{B})-\mathbf{W_{i\text{-}h}X}\|_F^2$

P4: $\arg\min_{\mathbf{Z}} \|\mathbf{X}-\mathbf{W_{h\text{-}o}Z}\|_F^2 + \lambda\|\mathbf{L}-\mathbf{DZ}\|_F^2 + \mu\|\mathbf{Z}-\phi(\mathbf{W_{i\text{-}h}X})-\mathbf{B}\|_F^2$

Sub-problems P1-P3 are simple least square minimizations. They have a closed form solution in the form of pseudo-inverse. They can also be solved efficiently using conjugate gradient. Sub-problem P4 is also a least square minimization problem; it becomes apparent after re-arranging, as follows

$$\arg\min_{\mathbf{Z}} \left\| \begin{pmatrix} \mathbf{X} \\ \sqrt{\lambda}\mathbf{L} \\ \sqrt{\mu}(\phi(\mathbf{W_{i\text{-}h}X})+\mathbf{B}) \end{pmatrix} - \begin{pmatrix} \mathbf{W_{h\text{-}o}} \\ \sqrt{\lambda}\mathbf{D} \\ \sqrt{\mu}\mathbf{I} \end{pmatrix} \mathbf{Z} \right\|_F^2 \quad (8)$$

Therefore in every iteration, we only need to solve sub-problems P1-P4; all of which have closed form solution. The final step in each iteration is to update the relaxation variable (by gradient descent).

$$\mathbf{B} \leftarrow \mathbf{Z}-\phi(\mathbf{W_{i\text{-}h}X})-\mathbf{B} \quad (9)$$

There are two exit criteria. The iterations continue till either a specified maximum number of iterations or till the difference between the objective function falls below a chosen threshold in successive iterations.

We construct a stacked architecture by nesting one discriminative AE inside the other. We learn all the layers in a greedy fashion, there is no fine-tuning stage.

# 3. Experimental Results

## 3.1. Description of Dataset

We demonstrate the performance of our proposed discriminative AE (DiAE) for the task of character recognition on MNIST digit dataset and its variants [31], USPS digit dataset [32], and Bangla and Devnagari [33] character datasets. The details of the datasets are given in table I and few sample images are shown in Fig 2.

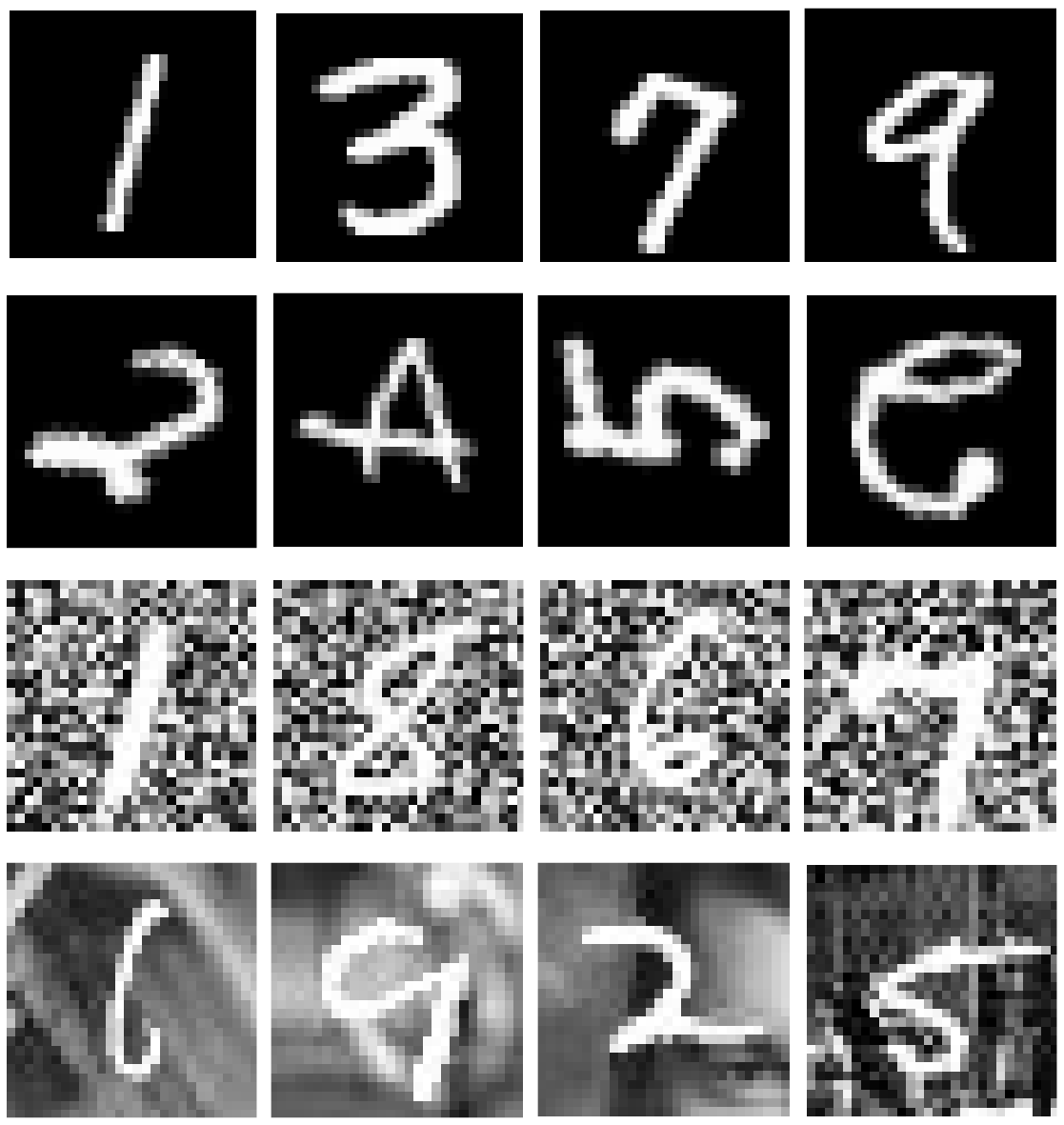

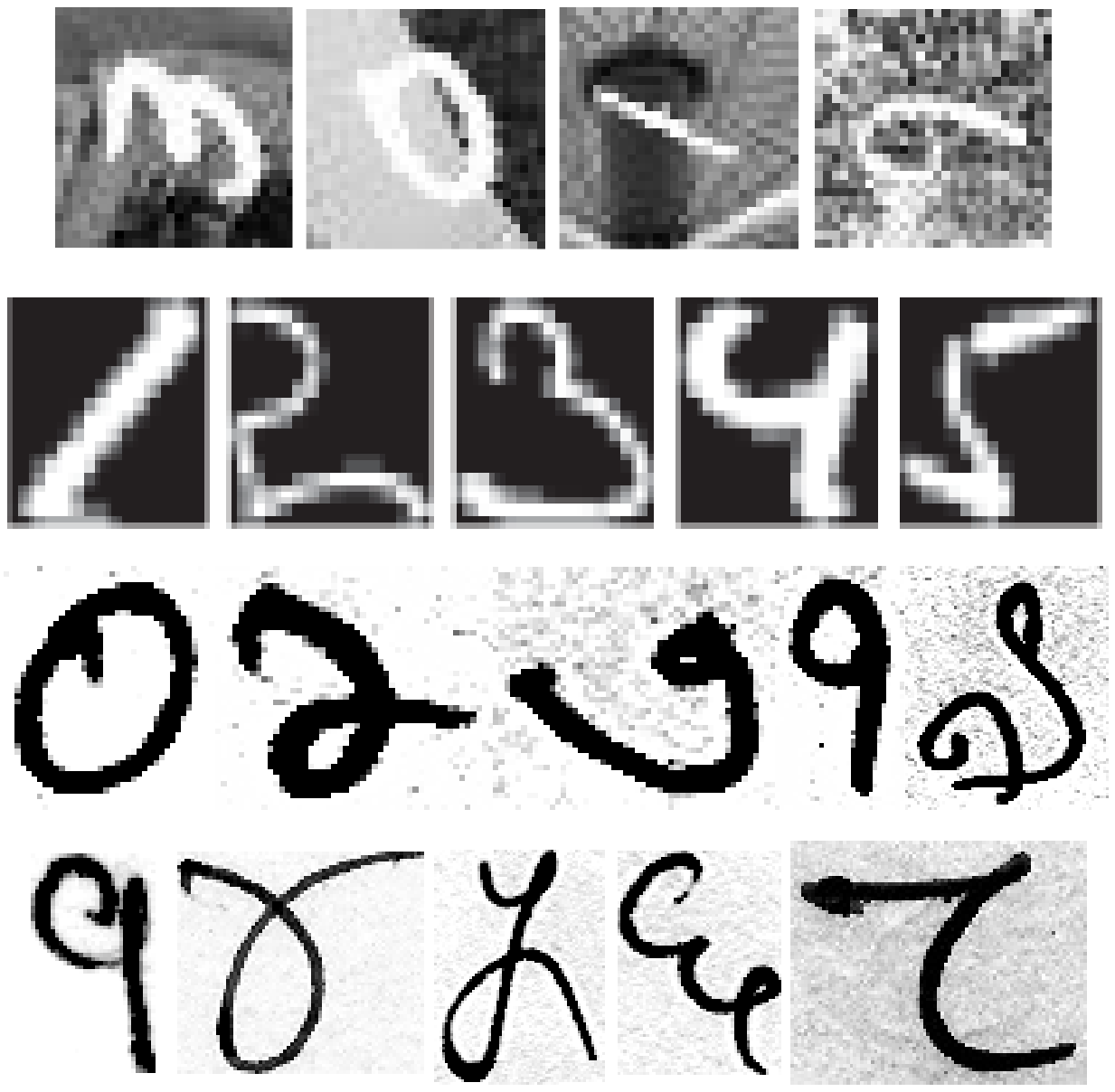

Fig. 2. Top to Bottom. MNIST Variations - basic, basic-rot, bg-rand, bg-img, bg-img-rot, USPS, Bangla and Devanagri

TABLE I
DATASET STATISTICS

| Dataset | Training/ Test Sample size | Description with (Dimension of each image sample) |
|---|---|---|
| MNIST | 60000/10000 | Basic digit dataset (28X28) |
| MNIST-Rot | 12000/50000 | Rotated digits (28X28) |
| MNIST-Back | 12000/50000 | Digits with background images (28X28) |
| MNIST-Rand | 12000/50000 | Digits with random background (28X28) |
| MNIST-RotBack | 12000/50000 | Rotated digits with background images (28X28) |
| USPS | 7291/2007 | Digitals from US postal Code (16X16) |
| Bangla | 19392/4000 | Bangla characters (32X32) |
| Devnagari | 18783/3763 | Devnagari characters (32X32) |

## 3.2. Experimental Setup

We used our DiAE structure to construct a 3-layer network with the hidden layer of the final stage forming the final feature set. The value of regularization parameter for each layer is computed using l-curve technique [34] to optimally recover the weights for each stage; the optimum value of lambda (regularization parameter) set as 1e+1 for each layer.

In each stage, the dimension of the feature set was reduced; optimum number of hidden layer variables is computed empirically. For MNIST dataset and its variations, the number of hidden layer nodes were 392-196-98; for USPS 230-200-170; for Bangla and Devnagari characters 900-700-500.

The feature set obtained using our design was fed into three classifiers - a parametric classifier – Multiclass SVM with RBF kernel and two non-parametric classifiers – KNN (K-nearest neighbor) with K=1 and SRC (Sparse Classifier) [35].

## 3.3 Experimental Results – Linear vs Nonlinear Activation Function

Most existing works on AE employ a non-linear activation function. However, recent studies [18, 19] advocated that the non-linear model essentially works in the linear region of the activation function only. Motivated by these results we empirically evaluate the performance of linear as well as non-linear activation functions.

Table II show the classification accuracy reported using KNN (K=1) classifier for MNIST dataset and its variants for standard AE (3) using both linear as well as non-linear activation function. A similar evaluation is shown for discriminative AE structure (4) in table III. For the standard AE design, we used the implementation provided in [36]. For the discriminative AE, we used our model with tanh (for non-linear activation) or Identity operator (for linear activation function).

Results reported in table II and III, show that linear activation function yields comparable or even better results than the non-linear model (sigmoid), at least for the character recognition task undertaken on MNIST benchmark datasets. Supported by these results, we report the results for our discriminative AE structure using linear activation function only in further sections.

TABLE II
CLASSIFICATION ACCURACY: STANDARD AE (LINEAR VS NON LINEAR)

| Dataset | Classification Accuracy (in %) | |
|---|---|---|
| | Linear Model | Non-Linear Model |
| MNIST | **96.33** | 96.11 |
| MNIST-Rot | **84.83** | 80.71 |
| MNIST-Back | **77.16** | 70.97 |
| MNIST-Rand | **86.42** | 81.11 |
| MNIST-RotBack | **52.21** | 44.6 |

TABLE III
CLASSIFICATION ACCURACY: DISCRIMINATIVE AE (LINEAR VS NON LINEAR)

| Dataset | Classification Accuracy (in %) | |
|---|---|---|
| | Linear Model | Non-Linear Model |
| MNIST | **97.15** | 96.98 |
| MNIST-Rot | **85.17** | 84.13 |
| MNIST-Back | **78.6** | 72.98 |
| MNIST-Rand | **87.61** | 85.99 |
| MNIST-RotBack | **53.5** | 45.45 |

## 3.4 Experimental Results – Proposed Approach vs Existing Models

We show the comparison of our approach – Discriminative AE with linear activation function – with two existing deep feature extraction techniques class sparse autoencoder (CSSAE) [36] and discriminative deep belief net (DDBN) [37] – these are relatively recent techniques in supervised representation learning; CSSAE has been proposed this year and DDBN is slightly old and has been used as a benchmark. Both these techniques (CCSAE and DDBN) have shown to outperform their unsupervised counterparts. The results for all three algorithms with different classifiers is reported in table IV.

For our proposed technique one needs to specify one parameter λ and one hyper-parameter μ. The parameter controls the relative importance of the autoencoder and discriminative penalties. Since there is no reason to favor one over the other we keep λ=1. Usually the hyper-parameter for augmented Lagrangian techniques need to be tuned. But for our problem it has a clear meaning. It controls the relative importance of the encoder and decoder terms. Since both are equally important we assign μ=1.

TABLE IV
CLASSIFICATION ACCURACY (IN %)

| | Using KNN | | | Using SRC | | | Using SVM | | |
|---|---|---|---|---|---|---|---|---|---|
| | Discriminative AE | DDBN | CSSAE | Discriminative AE | DDBN | CSSAE | Discriminative AE | DDBN | CSSAE |
| MNIST | **97.15** | 97.11 | 97.01 | **98.20** | 96.23 | 97.59 | **98.22** | 97.44 | 97.77 |
| MNIST-Rot | **85.17** | 84.51 | 82.47 | **90.21** | 89.21 | 87.76 | **86.53** | 83.50 | 82.06 |
| MNIST-Back | **78.6** | 77.06 | 73.91 | **85.4** | 80.19 | 79.91 | **85.59** | 80.29 | 77.91 |
| MNIST-Rand | **87.61** | 86.52 | 84.29 | **92.07** | 89.41 | 87.89 | **90.34** | 88.47 | 86.42 |
| MNIST-RotBack | **53.5** | 52.74 | 52.68 | **63.77** | 59.08 | 55.52 | **59.05** | 58.06 | 52.11 |
| USPS | **95.35** | 94.97 | 91.24 | **95.71** | 94.41 | 93.29 | **95.37** | 87.53 | 92.74 |
| Bangla | **86.70** | 85.99 | 82.98 | **92.75** | 92.59 | 90.65 | **84.00** | 83.21 | 82.79 |
| Devnagari | **93.05** | 92.36 | 92.03 | **96.07** | 95.33 | 94.94 | **87.14** | 87.08 | 84.07 |

It can be seen from the results that our design performs better that CSSAE or DDBN for all the datasets. For cases with inherent variations causing difficulty in classification such as MNIST-RotBack our model provides higher percentage improvement over DBN and SAE as compared to easier cases like MNIST. This highlights the effect of our discriminative design in mitigating the influence of unwanted image variations.

We also report the run times of all the three algorithms (coded in MATLAB) for the MNIST dataset in table V. The simulations are carried out on a machine with i7 CPU @ 3.10 GHz with 8 GB RAM. Our algorithm, based on MM technique, gives much smaller run times (~2.5 times faster than closest) than the other two compared against. Thus, our design of a stacked discriminative autoencoder not just improves accuracy but also enjoys the benefit of faster computations. This can be attributed to the use of linear activation function coupled with an efficient algorithm which achieves faster convergence (20 iterations).

TABLE V
RUN TIMES

| **Algorithm** | **DiAE** | **SAE** | **DBN** |
|---|---|---|---|
| **Run Time (sec.)** | 1948 | 18902 | 4916 |

## 3.4 Experimental Results – Analysis of our Discriminative Design

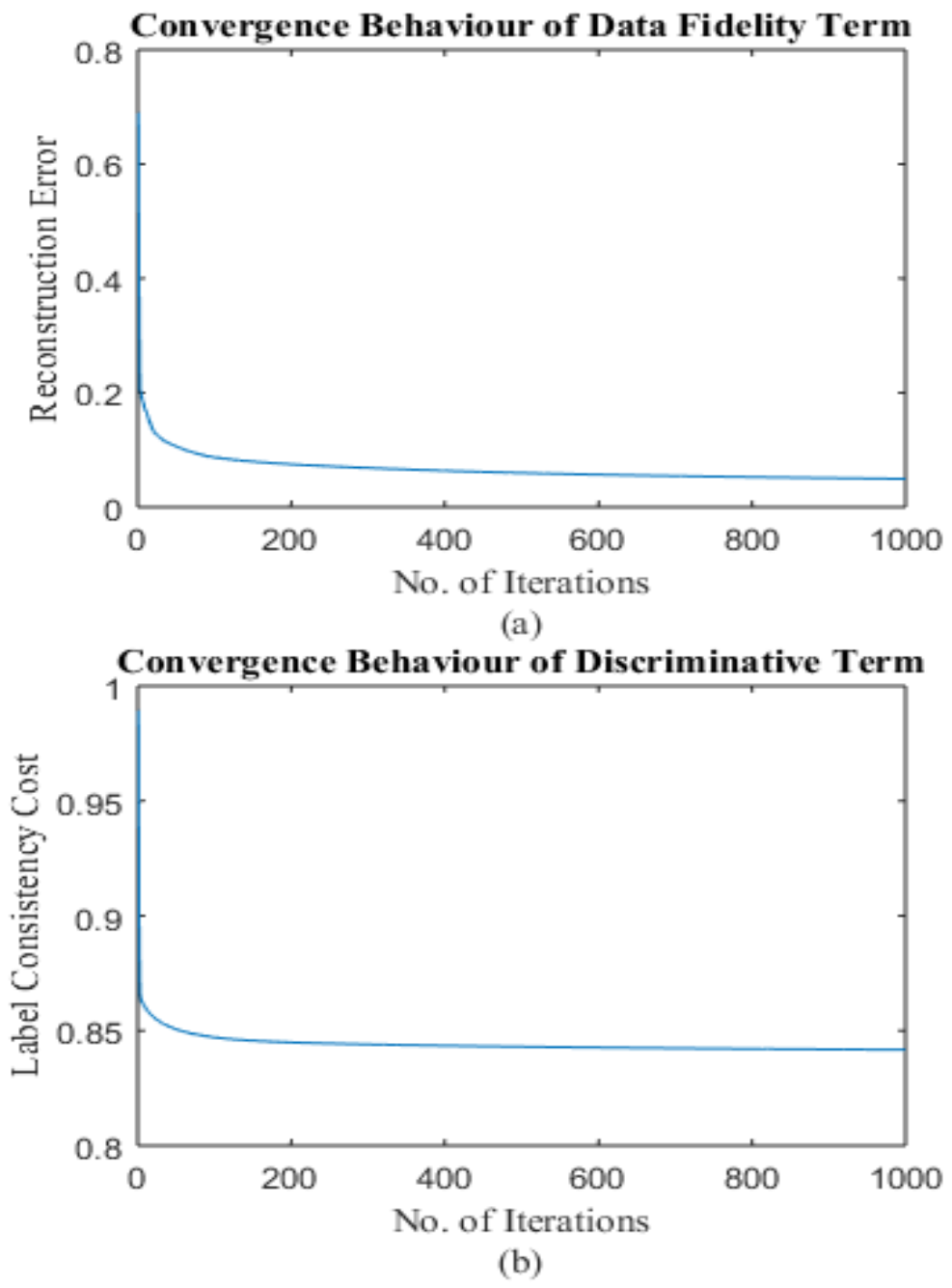

Fig. 3. Convergence Behavior of our Discriminative Autoencoder

In this section, we demonstrate the convergence behavior and discriminative capability of our model, DiAE. Figure 3 shows the convergence behavior of the algorithm for the MNIST-Rand dataset. Fig 3(a) indicates that as iterations proceed, the reconstruction loss of our proposed model reduces until convergence; ensuring data fidelity between the input and the derived feature representation. Fig. 3(b) shows the convergence of discriminative penalty term; indicating that over time our model is able to generate features consistent with the available class information.

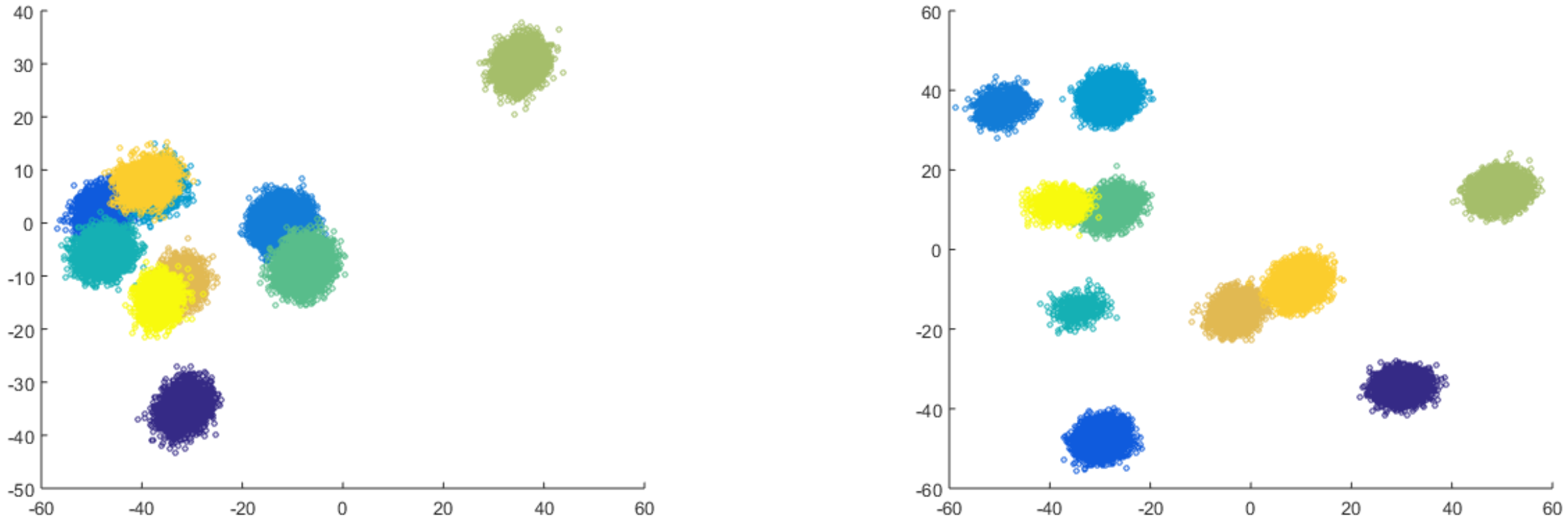

(a) (b)

Fig. 4. (a) Features from unsupervised autoencoder. (b) Features from proposed autoencoder

Figure 4 shows the feature vectors after dimensionality reduction by T-SNE. Fig. 4(a) shows an unsupervised one and Fig. 4(b) shows a supervised one. As can be seen, in our supervised version, the data is much well separated.

The code for our design is available at http://in.mathworks.com/matlabcentral/fileexchange/57347-discriminative-autoencder

# 4. Conclusion

In this work, we proposed a discriminative autoencoder framework. Our design involves linear mapping, instead of commonly employed non-linear activation, between the hidden layer and input/output layers. This considerably reduces the computational burden.

To improve the robustness of our model and alleviate the impact of image variations, we include additional discriminative (label consistent) constraints to the standard autoencoder Euclidean loss function. Use of label consistency term warrants that the feature vectors (hidden layer neurons) are recovered such that they are discriminant. Owing to this, all feature vectors corresponding to same class label share high similarity while the inter-class variability or discrimination increases.

We stack together our supervised AEs to form a stacked network useful for robust feature extraction. Experiments conducted on digit classification datasets validate our claim that our model yields higher classification accuracy than standard AE design with lower computational costs.